\documentclass[11pt,a4paper]{article}
\usepackage[T1]{fontenc}
\usepackage[utf8]{inputenc}
\usepackage{lmodern}
\usepackage[margin=25mm]{geometry}
\usepackage{amsmath,amssymb,booktabs,array,longtable,tabularx}
\usepackage{graphicx,xcolor,microtype,xurl}
\usepackage[numbers,sort&compress]{natbib}
\usepackage[colorlinks=true,linkcolor=blue!45!black,citecolor=blue!45!black,urlcolor=blue!45!black]{hyperref}
\hypersetup{pdftitle={GatedV: Self-Gated Values for Transformer Language Models},pdfauthor={Li-Ming Yu and Hongyu Zhou},pdfsubject={GatedV},pdfkeywords={Transformer, value projection, SiLU, paired experiments}}
\usepackage{enumitem,placeins}
\setlist{nosep,leftmargin=*}
\newcommand{\silu}{\operatorname{SiLU}}
\newcommand{\rms}{\operatorname{RMSNorm}}
\newcommand{\softmax}{\operatorname{softmax}}

\title{GatedV: Self-Gated Values for Transformer Language Models}
\author{
  Li-Ming Yu\thanks{Corresponding author. ORCID~\href{https://orcid.org/0000-0001-7552-8895}{0000-0001-7552-8895}.}\\
  Independent Researcher\\
  {\small\href{mailto:yuliming1996@gmail.com}{\texttt{yuliming1996@gmail.com}}}
  \and
  Hongyu Zhou\\
  Nanyang Technological University, Singapore\\
  {\small\href{mailto:hongyu006@e.ntu.edu.sg}{\texttt{hongyu006@e.ntu.edu.sg}}}
}

\begin{document}
\maketitle
\begin{abstract}

Standard attention forms query (Q), key (K), and value (V) matrices through learned linear projections of input hidden states. Architectures such as GAU and Mega incorporate nonlinear transformations into attention. In this paper, we propose GatedV, a simple attention variant that applies SiLU only to V before aggregation without introducing additional trainable parameters. We evaluate GatedV against a standard decoder-only Transformer using matched parameter counts, paired initializations, and equal training-token budgets. At 2.33M parameters, GatedV reduces mean held-out perplexity from 59.1 to 57.9 (2.0\%), with improvements across all five paired seeds. At 26.16M parameters, following an equal-budget learning-rate search, mean held-out perplexity falls from 20.2 to 19.7 (2.4\%), with improvements across all three paired seeds. Small-model ablations show that the tested affine rescaling of values does not fully reproduce GatedV's gains. A short interleaved benchmark on NVIDIA DGX Spark measures approximately 0.7\% higher full-context forward latency for the larger model, based on synchronized wall-clock timings. These findings suggest that value-only self-gating can improve language modeling performance without increasing parameter count or redesigning the attention-weighting mechanism.
\end{abstract}

\section{Introduction}

Standard Transformer blocks combine self-attention with position-wise feed-forward networks~\citep{transformer}. The feed-forward sublayer applies nonlinear transformations to individual token representations, whereas attention mixes information across tokens. Within attention, queries (Q) and keys (K) determine the mixing weights, and values (V) provide the representations to be combined. These three branches are usually formed by linear projections, despite their different roles. An additional nonlinearity on queries or keys changes the weighting calculation; on values, it changes what enters the weighted sum. Moreover, applying a nonlinearity before aggregation is generally not equivalent to applying it afterward.

Nonlinear value representations are already used in several attention architectures. Mega applies SiLU to projected values in a block that also uses moving-average features and additional gates~\citep{mega}. HYLA introduces nonlinear value computation without additional parameters, together with normalization across attention heads~\citep{hypernetwork}. Neural Attention instead replaces the linear Q/K/V projections with multilayer perceptrons~\citep{neuralattention}. Gated Attention compares gates at several locations, including the value projection, with its main results using a learned sigmoid gate after scaled dot-product attention~\citep{gatedattention}. GLU Attention introduces separate content and gate projections and reduces the aggregated value width to match the combined value/output projection-weight count~\citep{gluattention}. These approaches differ in both the source of the nonlinearity and the computation surrounding it.

In this paper, we ask whether a nonlinear value transform improves a standard Transformer when the other components are left unchanged. A parameter-free elementwise activation allows the original projection dimensions and block configuration to be retained, while altering only the operation applied to the projected values.
We call this variant GatedV (self-gated values). It applies SiLU to the usual value projection $V$ before attention aggregation:
\[
    \widetilde V=\silu(V)=V\odot\sigma(V),
\]
where $\sigma$ is the elementwise sigmoid function. The same projected values supply both the content and its gate, so no separate learned gate projection is needed. The query/key projections, softmax weighting rule, and remaining Transformer components are retained. GatedV therefore has exactly the same number of trainable parameters as the baseline, without adjusting the depth or width of either model.

We train baseline and GatedV decoder-only language models from scratch on FineWeb-Edu~\citep{fineweb}, at 2.33M and 26.16M parameters and with equal training-token budgets within each scale. GatedV reduces mean held-out perplexity by 2.0\% and 2.4\%, respectively, with improvements in every paired confirmation run. Additional experiments examine activation placement, learning-rate sensitivity, initial value statistics, and computational cost. Whether these improvements persist at larger scales, on other corpora, or on tasks beyond next-token prediction remains to be tested.
Code, experiment recipes, and recorded results are available at \url{https://github.com/yuliming1996/GatedV}.

\section{Background and model formulation}
\label{sec:background}

\subsection{Transformer backbone}
\label{sec:backbone}
Our baseline is a decoder-only Transformer with causal multihead attention~\citep{transformer,gpt}. Each block uses RMSNorm before its attention and feed-forward sublayers~\citep{rmsnorm}, rotary positional embeddings for queries and keys~\citep{rope}, and a SwiGLU feed-forward transformation~\citep{swiglu}.

Let $X\in\mathbb{R}^{T\times d}$ denote the hidden states entering a block, where $T$ is the sequence length and $d$ is the model width. We omit the batch dimension and use row-vector notation. The attention sublayer first computes
\begin{equation}
 U=\rms_{\gamma_a}(X),\qquad
 Q=UW_Q,\quad K=UW_K,\quad V=UW_V,
 \label{eq:projections}
\end{equation}
where $W_Q,W_K,W_V\in\mathbb{R}^{d\times d}$. RMSNorm acts on each token independently:
\begin{equation}
 \rms_\gamma(x)=\gamma\odot
 \frac{x}{\sqrt{d^{-1}\sum_{j=1}^{d}x_j^2+\epsilon}},
 \qquad \epsilon=10^{-6}.
 \label{eq:rmsnorm}
\end{equation}
The projected tensors are split into $H$ heads of width $d_h=d/H$. For head $h$, let $\widehat Q_h=\operatorname{RoPE}(Q_h)$ and $\widehat K_h=\operatorname{RoPE}(K_h)$. We use a rotary base of 10,000. The attention weights and output are
\begin{equation}
 A_h=\softmax\!\left(
 \frac{\widehat Q_h\widehat K_h^\top}{\sqrt{d_h}}+M
 \right),\qquad
 O=\operatorname{Concat}_{h=1}^{H}(A_hV_h),
 \label{eq:attention}
\end{equation}
where $M_{ij}=0$ for $j\leq i$ and $M_{ij}=-\infty$ otherwise. Softmax is applied across the key positions in each row. Values enter this aggregation linearly for fixed $A_h$, although attention as a function of $X$ is nonlinear because the weights themselves depend on the input.

The block then applies the two residual updates
\begin{align}
 X'&=X+OW_O,\\
 Z&=\rms_{\gamma_f}(X'),\\
 X_{\mathrm{next}}&=X'+
 \big[\silu(ZW_1)\odot(ZW_2)\big]W_3,
 \label{eq:block}
\end{align}
where $W_O\in\mathbb{R}^{d\times d}$, $W_1,W_2\in\mathbb{R}^{d\times d_{\mathrm{ff}}}$, and $W_3\in\mathbb{R}^{d_{\mathrm{ff}}\times d}$. The SiLU activation is $\silu(z)=z\sigma(z)$, with $\sigma(z)=(1+e^{-z})^{-1}$~\citep{silu,swish}; both the activation and $\odot$ act elementwise. After $L$ blocks, a final RMSNorm precedes the vocabulary readout, whose weights are tied to the input embedding. All learned linear maps are bias-free. Dropout and additional headwise query/key normalization are disabled.

\subsection{Nonlinear projections and attention gates}
\label{sec:related}
Nonlinear features can be introduced at several stages of attention, beginning with the projections that construct Q, K, and V. Neural Attention replaces these linear maps with two-layer networks containing LayerNorm and ReLU~\citep{neuralattention}. This increases the projection parameter count. Its ReLU ablation retains the expanded key/value projections and LayerNorm, testing the activation within the modified network rather than comparing with a parameter-matched standard Transformer.

Nonlinear values also appear in architectures that redesign the attention block more broadly. The Gated Attention Unit (GAU) combines nonlinear values and shared query/key features with squared-ReLU attention and multiplicative gating in a unified attention and feed-forward layer~\citep{gau}. Mega builds on this construction, deriving shared query/key features from exponential moving averages and adding reset and update gates~\citep{mega}. Its value branch applies SiLU directly to the projected inputs before aggregation. This provides a direct precedent for value-side SiLU, although the surrounding computation differs from conventional multihead attention.

GLU Attention retains softmax aggregation and uses two independently projected branches, $\widetilde V=(UW_C)\odot\silu(UW_G)$~\citep{gluattention}. For model width $d$, it sets the combined value width across heads to $m=2d/3$. Ignoring biases, the two $d\times m$ input projections and the $m\times d$ output projection then contain $3dm=2d^2$ weights, matching the standard value and output projections. Thus its parameter matching involves narrower values and a separate learned gate. GatedV instead preserves the original value width and projection matrices, applying SiLU directly to a single projected value tensor.

Other formulations make the nonlinear transformation depend on how values are combined. HYLA applies ReLU to an attention-weighted combination of value features across heads for each query--key pair. It introduces no additional parameters and uses RMS normalization across heads instead of key-wise softmax~\citep{hypernetwork}. LASER retains softmax weights but combines an exponential value transform with a logarithm after aggregation, giving $\log\!\big(A_h\exp(V_h)\big)$~\citep{laser}. These operations are not generally equivalent to applying an activation independently to each projected value and then taking the usual weighted sum.

Gate placement has also been studied while retaining softmax attention. Gated Attention compares learned gates on Q/K/V and at the attention output, with its main configuration using a head-specific sigmoid gate after scaled dot-product attention~\citep{gatedattention}. Its G2 gate uses a separate input projection to compute a sigmoid gate on the values before aggregation. It also evaluates SiLU applied directly to the attention output without additional parameters (their Table~3, row~6). Two distinctions are relevant here: whether the gate uses a separate learned projection, and whether it acts before or after aggregation. We examine self-gating of the projected values before aggregation, retaining the original projections and the softmax weighting rule.

\subsection{GatedV}
\label{sec:model}
GatedV replaces the projected values in Equation~\ref{eq:projections} with
\begin{equation}
 \widetilde V=\phi(V)=V\odot\sigma(V),
 \qquad \phi=\silu.
 \label{eq:selfgate}
\end{equation}

\begin{figure}[t]
\centering
\includegraphics[width=\linewidth]{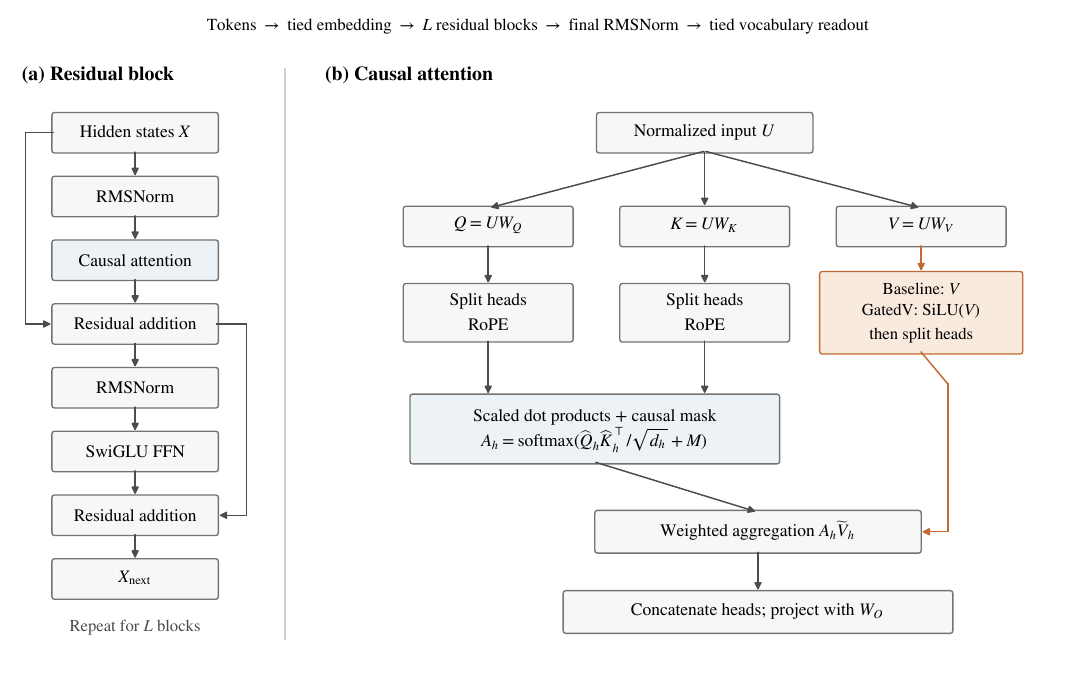}
\caption{Transformer block and GatedV value transformation. The baseline aggregates the projected values $V$, whereas GatedV applies $\silu(V)$ before aggregation. The value box shows the two alternatives; each model uses one.}
\label{fig:architecture}
\end{figure}

The Q/K/V projections are implemented as one fused linear map. After splitting its output into three tensors, we apply $\phi$ to $V$ before reshaping into heads. The per-head comparison is therefore
\begin{equation}
 Y_h^{\mathrm{base}}=A_hV_h,\qquad
 Y_h^{\text{GatedV}}=A_h\phi(V_h).
 \label{eq:vonly}
\end{equation}
Figure~\ref{fig:architecture} shows where the value activation is applied.

The gate is determined by the same projected value that it multiplies. It requires no separate projection such as $W_G$ in $(UW_V)\odot\sigma(UW_G)$, the two-projection form of a gated linear unit~\citep{glu}. Placement also matters: in general,
\begin{equation}
 A_h\phi(V_h)\neq\phi(A_hV_h).
 \label{eq:placement}
\end{equation}
The left-hand side transforms each source token's value before mixing it with other values; the right-hand side transforms the resulting mixture.

For a fixed normalized input $U$ and fixed query/key projection weights, GatedV leaves $A_h$ unchanged. No trainable tensors are added or resized. The depth, hidden widths, head structure, normalization, positional encoding, and feed-forward blocks remain as defined above, so paired baseline and GatedV models can start from identical weights. The activation adds $O(Td)$ elementwise work per block. 

For the activation-placement experiments, we also apply $\phi$ to queries alone or to all three of Q, K, and V. Query and key activations are applied before RoPE. These variants retain the original depth and projection dimensions.

\FloatBarrier

\section{Experimental design}
\label{sec:experiments}

\subsection{Models, data, and training}
\label{sec:training}
We compare baseline and GatedV models at two sizes (Table~\ref{tab:backbone}), with identical initial weights and sampled training windows. Within each pair, only the value activation differs.

\begin{table}[t]
\centering\small
\caption{Main confirmation settings. Both models use a vocabulary of 8,192, context length 256, and batch size 8. Target exposures count sampled prediction targets, including repetitions.}
\label{tab:backbone}
\begin{tabular}{lrr}
\toprule
Setting & Small & Medium\\
\midrule
Layers / model width & 6 / 128 & 12 / 384\\
Heads / head width & 4 / 32 & 6 / 64\\
SwiGLU hidden width & 384 & 1,152\\
Trainable parameters & 2,328,192 & 26,158,464\\
Training-corpus tokens & 40,001,310 & 512,000,068\\
Training-corpus documents & 27,830 & 212,052\\
Updates per run & 32,768 & 327,680\\
Target exposures per run & 67,108,864 & 671,088,640\\
Warmup updates & 512 & 5,120\\
Confirmation peak learning rate & 0.0048 & 0.0003\\
Paired confirmation seeds & 5 & 3\\
\bottomrule
\end{tabular}
\end{table}

Training uses FineWeb-Edu text~\citep{fineweb} and a fixed 8,192-entry byte-level BPE tokenizer~\citep{gpt2} trained on WikiText training data~\citep{wikitext}. Each training sequence is sampled from a single document and contains 256 input tokens with their next-token targets. The small confirmation held-out set and the medium development and held-out sets each contain 2,048 documents from distinct domain groups, with the first eligible 256-target window per document.

We train with AdamW~\citep{adamw}, using $(\beta_1,\beta_2)=(0.9,0.95)$, $\epsilon=10^{-8}$, matrix weight decay 0.1, no decay on normalization gains, and gradient-norm clipping at 1. After warmup, the learning rate follows a cosine schedule to 10\% of its peak. We evaluate final fixed-budget checkpoints. Training uses FP32. We initialize projection and embedding matrices from a Gaussian distribution with fan-in scaling, with an additional factor of $1/\sqrt{2L}$ for attention output and feed-forward down-projections, where $L$ is the number of Transformer layers.

\subsection{Learning-rate selection and controls}
The small-model confirmation compares baseline and GatedV at a shared peak learning rate of 0.0048 over five paired seeds. We fixed this rate after a 4,096-update development screen in which both models achieved their lowest NLL at the highest rate tested.

After the original medium-model confirmation at 0.0006, we select each architecture's learning rate from $\{0.00001,0.0001,0.0003,0.0006\}$ by the lowest final development NLL, using one shared selection seed (20270201) and equal training budgets; ties favor the lower rate. Both select 0.0003 (Table~\ref{tab:medium_lr}). We evaluate these configurations with the original three paired seeds (20270211--20270213), distinct from the selection seed, on the previously used medium held-out set. The original 0.0006 confirmation is retained in Appendix~\ref{sec:original_medium}. A separate development-set comparison evaluates value gates at $\{0.0006,0.0003,0.0001\}$ (Section~\ref{sec:gate_comparison}).

\begin{table}[htbp]
\centering\small
\caption{Medium-model learning-rate selection: final development PPL at the shared selection seed 20270201, with 327,680 updates per run. Both architectures select 0.0003.}
\label{tab:medium_lr}
\begin{tabular}{lrrrr}
\toprule
Variant & LR 0.00001 & LR 0.0001 & LR 0.0003 & LR 0.0006\\
\midrule
Baseline & 36.853 & 20.938 & 20.558 & 21.558\\
GatedV & 37.451 & 20.632 & 20.062 & 21.005\\
\bottomrule
\end{tabular}
\end{table}

Small-model ablations compare activation placement and learning rates using three paired seeds and 32,768 updates per run (Table~\ref{tab:ablations}). We also test whether fixed adjustments to the initial value statistics can reproduce GatedV's gain. Affine match rescales and shifts the baseline values to match the mean and variance of SiLU-transformed values; Moment-matched SiLU applies the reverse adjustment to SiLU-transformed values, matching the original value statistics. For each seed and layer, we estimate $(\mu_V,\sigma_V)$ and $(\mu_\phi,\sigma_\phi)$ from 128 training windows at baseline initialization and define
\begin{align}
 V_{\mathrm{aff}}&=\frac{\sigma_\phi}{\sigma_V}(V-\mu_V)+\mu_\phi,\\
 V_{\mathrm{mm}}&=\frac{\sigma_V}{\sigma_\phi}\big(\phi(V)-\mu_\phi\big)+\mu_V.
 \label{eq:controls}
\end{align}
The coefficients remain fixed throughout training. At peak learning rates 0.0024 and 0.0048, with three paired seeds and 32,768 updates per run, both controls have higher mean development PPL than GatedV at the matching rate, by 1.77--2.82\% (see Table~\ref{tab:moment} and Section~\ref{sec:activation_LR}). Thus these fixed moment adjustments do not reproduce GatedV's full gain.

\subsection{Evaluation and uncertainty}
For seed $s$, document loss sums $\ell_{s,i}$, and target counts $n_i$, we report token-averaged negative log-likelihood and perplexity as
\begin{equation}
 L_s=\frac{\sum_i\ell_{s,i}}{\sum_i n_i},\qquad P_s=\exp(L_s),\qquad
 \Delta_P=100\left(\frac{\overline P_{\mathrm{GatedV}}}{\overline P_{\mathrm{base}}}-1\right).
 \label{eq:pplchange}
\end{equation}
$\overline P$ is the arithmetic mean of per-seed perplexities, not the exponential of mean NLL. Negative $\Delta_P$ denotes improvement. For multi-seed PPL results, $\pm$ denotes the sample standard deviation across training seeds.

To estimate uncertainty from evaluation texts in the main held-out comparisons~\citep{benchvariance,statsguide}, we keep the trained models fixed and average each document's summed loss difference across seed pairs. We then draw 2,000 bootstrap samples of the documents, with replacement and at the original sample size. For each sample, we recompute the token-averaged NLL difference; the 2.5th and 97.5th percentiles of these estimates define the 95\% bootstrap interval (see Section~\ref{sec:heldout}).

\section{Results}
\label{sec:results}

\subsection{Held-out language modeling}
\label{sec:heldout}
Table~\ref{tab:confirmation} summarizes the two confirmations. At 2.33M parameters, mean held-out PPL decreases by 2.03\%, with all five pairs improving. Individual reductions range from 0.33\% to 3.74\%. At 26.16M parameters and the development-selected learning rate of 0.0003, all three confirmation pairs improve, reducing mean held-out PPL by 2.40\%. Individual reductions range from 2.22\% to 2.66\%.

\begin{table}[htbp]
\centering\small
\caption{Held-out confirmation results. Values are arithmetic mean PPL $\pm$ sample SD. The studies use different held-out sets and training configurations; the medium model uses peak LR 0.0003.}
\label{tab:confirmation}
\begin{tabular}{lrrrr}
\toprule
Model size & Baseline & GatedV & $\Delta_P$ & Improved pairs\\
\midrule
2.33M & $59.118\pm0.530$ & $57.916\pm0.782$ & $-2.03\%$ & 5/5\\
26.16M & $20.233\pm0.068$ & $19.748\pm0.047$ & $-2.40\%$ & 3/3\\
\bottomrule
\end{tabular}
\end{table}

The mean GatedV-minus-baseline NLL differences are $-0.020591$ and $-0.024288$ for the small and medium models, respectively. The corresponding 95\% document-bootstrap intervals, with the trained models held fixed, are $[-0.021869,-0.019356]$ and $[-0.025507,-0.023153]$. After averaging NLL across the three seeds for each model, we find that GatedV has lower NLL on 1,672 of the 2,048 medium-model held-out windows (81.6\%).

\begin{figure}[t]
\centering
\includegraphics[width=\linewidth]{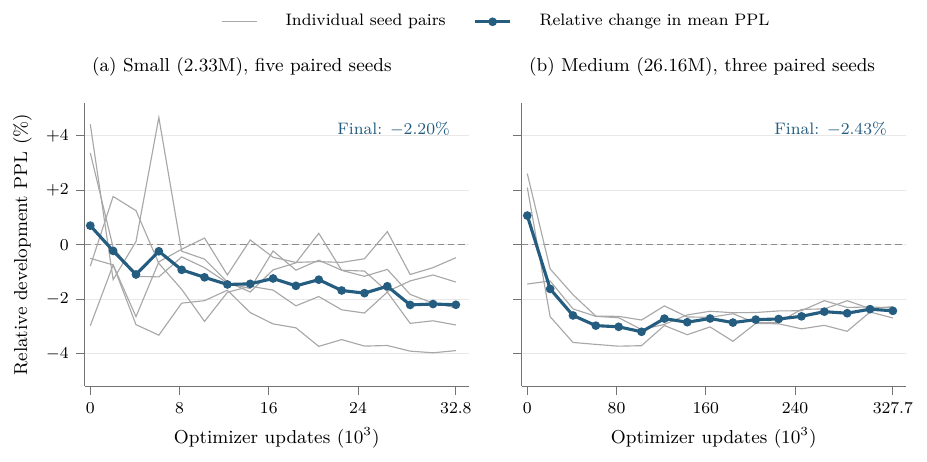}
\caption{Relative development PPL, $\Delta\mathrm{PPL}/P_{\mathrm{base}}$, shown as a percentage; $\Delta\mathrm{PPL}$ denotes GatedV minus baseline. Thin gray curves show individual seed pairs; the thick blue curve uses the arithmetic mean PPL of each model across seeds. Negative values favor GatedV.}
\label{fig:development_delta_ppl}
\end{figure}

\begin{figure}[t]
\centering
\includegraphics[width=\linewidth]{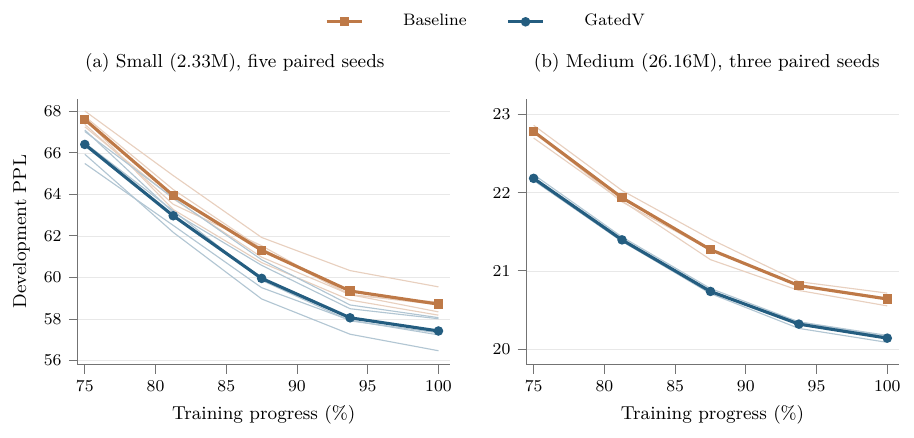}
\caption{Development PPL over the final 25\% of training. Copper denotes baseline and blue denotes GatedV. Thin light curves show individual runs; thick curves show arithmetic means across seeds. Held-out endpoints are reported separately in Table~\ref{tab:confirmation}.}
\label{fig:development_ppl}
\end{figure}

Figure~\ref{fig:development_delta_ppl} shows the relative change in development PPL throughout training, including initialization. At each recorded update, we exponentiate each run's development NLL and then average PPL across seeds; the mean curve uses the ratio of these arithmetic means, as in Equation~\ref{eq:pplchange}. The positive initial $\Delta_P$ in both panels may reflect the different activation statistics produced by SiLU under the shared Gaussian fan-in initialization. Figure~\ref{fig:development_ppl} shows absolute PPL over the final quarter of training. Final development means are 58.716 versus 57.422 for the small model and 20.640 versus 20.139 for the medium model, corresponding to reductions of 2.20\% and 2.43\%. 

\subsection{Activation placement and learning rate}
\label{sec:activation_LR}
Table~\ref{tab:ablations} summarizes the small-model activation-placement and learning-rate comparisons. At the same six-layer depth, value-only SiLU improves all three pairs. Query-only SiLU worsens all three, while joint Q/K/V SiLU slightly lowers mean PPL, with improvements in two of the three pairs.

\begin{table}[t]
\centering\small
\caption{Development-set comparisons. The two groups of rows use different sets of three paired seeds. $\Delta_P$ compares the variant with the baseline.}
\label{tab:ablations}
\begin{tabular}{lrrrr}
\toprule
Variant or peak LR & Baseline PPL & Variant PPL & $\Delta_P$ & Improved pairs\\
\midrule
\multicolumn{5}{l}{\textrm{Activation placement; LR = 0.0048}}\\
Q-only & $58.580\pm0.557$ & $59.522\pm0.337$ & $+1.61\%$ & 0/3\\
GatedV & $58.580\pm0.557$ & $58.039\pm0.525$ & $-0.92\%$ & 3/3\\
All Q/K/V & $58.580\pm0.557$ & $58.458\pm0.485$ & $-0.21\%$ & 2/3\\
\midrule
\multicolumn{5}{l}{\textrm{Small-model LR study; variant is GatedV}}\\
0.0012 & $56.441\pm0.548$ & $55.269\pm1.120$ & $-2.08\%$ & 2/3\\
0.0024 & $56.180\pm0.462$ & $54.450\pm0.140$ & $-3.08\%$ & 3/3\\
0.0048 & $58.770\pm0.983$ & $57.384\pm0.168$ & $-2.36\%$ & 3/3\\
0.0096 & $65.168\pm1.333$ & $64.063\pm0.442$ & $-1.70\%$ & 2/3\\
\bottomrule
\end{tabular}
\end{table}

In the separate 32,768-update exploratory small-model learning-rate grid, both architectures select the interior rate 0.0024 by mean final development NLL. GatedV improves mean PPL by 3.08\% there and improves every pair. Its mean is favorable at all four rates, but one pair worsens at each of 0.0012 and 0.0096. The medium-model comparison at lower learning rates is reported in Table~\ref{tab:value_gates}.

Relative to baseline, Affine match changes mean development PPL by $-1.12\%$ at LR 0.0024 and $+0.26\%$ at 0.0048 (Table~\ref{tab:moment}). Moment-matched SiLU uses a fixed shift and rescaling to match SiLU outputs to the initial baseline value mean and variance; its PPL reductions are 0.35\% and 0.63\%, respectively. Both controls have higher PPL than GatedV in every pair.

\begin{table}[t]
\centering\small
\caption{Final development PPL for fixed moment controls, with three paired seeds. Relative changes compare each control with baseline or GatedV at the same peak LR.}
\label{tab:moment}
\begin{tabular}{lrrrr}
\toprule
Control & Peak LR & PPL $\pm$ SD & vs baseline & vs GatedV\\
\midrule
Affine match & 0.0024 & $55.551\pm0.498$ & $-1.12\%$ & $+2.02\%$\\
Moment-matched SiLU & 0.0024 & $55.983\pm0.787$ & $-0.35\%$ & $+2.82\%$\\
Affine match & 0.0048 & $58.920\pm0.298$ & $+0.26\%$ & $+2.68\%$\\
Moment-matched SiLU & 0.0048 & $58.399\pm0.540$ & $-0.63\%$ & $+1.77\%$\\
\bottomrule
\end{tabular}
\end{table}

\subsection{Computational cost}
We benchmark the six medium-model checkpoints from the 0.0006 confirmation on NVIDIA DGX Spark with fixed synthetic inputs, batch size 8, and context length 256. The test takes about 97 seconds on a shared host, alternating models over seven rounds with three warmups and ten timed calls per mode. Wall-clock timings use CUDA synchronization. We compute paired time ratios from median call times, take the median across rounds for each seed, and average across seeds.

Measured latency increases by 0.70\% for full-context forward passes and 1.14\% for training steps with GatedV (Table~\ref{tab:cost}). Across the 21 paired comparisons, the relative change in forward latency for GatedV versus baseline ranges from $-1.15\%$ to $+2.57\%$. Median peak CUDA allocated memory rises from 1.7525 to 1.8569 GiB (5.96\%).

\begin{table}[t]
\centering\small
\caption{Medium-model time ratios relative to baseline and median peak CUDA allocated memory over the forward and training-step measurements combined.}
\label{tab:cost}
\begin{tabular}{lrr}
\toprule
Measure & Baseline & GatedV\\
\midrule
Full-context forward time ratio & 1.000000 & 1.007003\\
Training-step time ratio & 1.000000 & 1.011433\\
CUDA peak allocated (GiB) & 1.7525 & 1.8569\\
\bottomrule
\end{tabular}
\end{table}

\section{Discussion}
\label{sec:analysis}
\label{sec:discussion}
GatedV improves held-out language modeling in both tested configurations without adding parameters. In the medium model, its advantage narrows from 3.66\% at peak LR 0.0006 to 2.40\% at the development-selected rate of 0.0003: lowering the rate improves the baseline more, while GatedV remains better in all three pairs (Appendix~\ref{sec:original_medium}). The branch comparisons show that this benefit is not shared by every activation placement.

\subsection{Comparison with GLU, tanh, and G2}
\label{sec:gate_comparison}
GatedV uses a single value projection with sigmoid self-gating. We compare it with two-projection GLU value gates, a parameter-free tanh alternative, and a learned headwise G2 gate in the medium model (Table~\ref{tab:value_gates}). The GLU controls use $(UW_C)\odot\silu(UW_G)$ before aggregation, following GLU Attention~\citep{gluattention}. The full-width control retains 384 value dimensions, adding a separate projection; the reduced-width control uses 258 dimensions, the nearest six-head-compatible total width to $2d/3$, to approximate the baseline parameter count. The tanh control applies $V\odot\tanh(V)$ at the original width. Headwise G2 applies $V_h\,\sigma((UW_G)_h)$ before aggregation, using one scalar sigmoid gate per token and head~\citep{gatedattention}. Its bias-free $384\times6$ gate projection in each of the 12 layers adds 27,648 parameters. All runs process the same sampled training windows and share identical initial weights for corresponding matrices with matching shapes.

\begin{table}[htbp]
\centering\small
\caption{Exploratory medium-model value-gate comparison: final development PPL with one shared seed (20270201), reused development data, and 327,680 updates per run.}
\label{tab:value_gates}
\begin{tabular}{lrrrr}
\toprule
Variant & Parameters (M) & LR 0.0006 & LR 0.0003 & LR 0.0001\\
\midrule
Baseline & 26.158 & 21.558 & 20.558 & 20.938\\
GatedV & 26.158 & 21.005 & 20.062 & 20.632\\
Headwise G2 & 26.186 & 21.568 & 20.630 & 21.189\\
Full-width GLU & 27.928 & 21.197 & 20.309 & 20.930\\
Reduced-width GLU & 26.186 & 21.522 & 20.657 & 21.382\\
$V\odot\tanh(V)$ & 26.158 & 21.337 & 20.325 & 20.986\\
\bottomrule
\end{tabular}
\end{table}

All six variants achieve their lowest final development PPL at the interior rate 0.0003. The separate GLU gate does not improve on GatedV in this grid, either at full width with more parameters or at reduced width with a similar parameter count. At the selected rate, the tanh control also improves on the baseline, but GatedV remains 1.29\% lower in PPL (2.42\% below the baseline). Headwise G2 has slightly higher PPL than baseline at all three rates; at 0.0003, GatedV is 2.76\% lower than G2. On this shared seed, GatedV has the lowest development PPL at each of the three learning rates.

\subsection{Limitations}
\label{sec:limitations}
Our experiments are limited to models below 28M parameters on FineWeb-Edu, using 256-token contexts and next-token prediction. Larger models, longer contexts, other corpora, and downstream tasks remain untested. The GLU, tanh, and G2 comparisons use a single seed, and cost measurements cover only our FP32 implementation.

\section{Conclusion}
\label{sec:conclusion}
Applying SiLU to values before aggregation improves held-out perplexity by 2.03\% and 2.40\% in the two tested Transformer configurations, with all paired confirmation runs improving. The change preserves parameter count, depth, and width. The exploratory controls show that the gain is not shared by every activation placement or learning rate, and that the tested initial-moment adjustments do not reproduce it fully. The measured runtime increase is small in this implementation, but its precise magnitude and the extension to larger models, other tasks, and optimized inference remain unresolved.

\appendix
\section{Medium-model learning-rate comparison}
\label{sec:original_medium}
The original medium-model search used $\{0.0006,0.0012,0.0024\}$ and selected its lower boundary, 0.0006, for both architectures. Table~\ref{tab:medium_lr_confirmation} retains that confirmation alongside the 0.0003 results. Both rates use the same three seed pairs, matched initial weights and training-window schedules, 327,680 updates per run, and the same development and held-out sets.

\begin{table}[htbp]
\centering\small
\caption{Medium-model confirmation at two peak learning rates. PPL is the arithmetic mean $\pm$ sample SD across seeds 20270211--20270213; $\Delta_P$ compares GatedV with baseline within each row.}
\label{tab:medium_lr_confirmation}
\begin{tabular}{llrrr}
\toprule
Evaluation set & Peak LR & Baseline PPL & GatedV PPL & $\Delta_P$\\
\midrule
Development & 0.0006 & $21.804\pm0.182$ & $21.039\pm0.045$ & $-3.51\%$\\
Development & 0.0003 & $20.640\pm0.082$ & $20.139\pm0.047$ & $-2.43\%$\\
\midrule
Held-out & 0.0006 & $21.384\pm0.166$ & $20.600\pm0.018$ & $-3.66\%$\\
Held-out & 0.0003 & $20.233\pm0.068$ & $19.748\pm0.047$ & $-2.40\%$\\
\bottomrule
\end{tabular}
\end{table}

Reducing the peak learning rate from 0.0006 to 0.0003 lowers mean held-out PPL by 5.38\% for baseline and 4.14\% for GatedV. All three pairs favor GatedV at each rate. We report the rates separately because they measure the advantage under different optimization settings.

\FloatBarrier
\section{Value activation during training and inference}
\label{sec:cross_eval}
We held the weights of each final medium-model checkpoint (peak LR 0.0003) fixed and evaluated it on the same 2,048 held-out windows with either linear values or SiLU applied to the values in every attention layer. Switching the value transformation at inference raised PPL for all six checkpoints, most sharply when we removed SiLU from GatedV (Table~\ref{tab:cross_eval}).

\begin{table}[htbp]
\centering\small
\caption{Held-out PPL (mean and sample standard deviation across three paired seeds).}
\label{tab:cross_eval}
\begin{tabular}{lrr}
\toprule
Training & Linear at inference & SiLU at inference\\
\midrule
Linear (baseline) & $20.233\pm0.068$ & $102.565\pm31.053$\\
SiLU (GatedV) & $4{,}300.232\pm899.037$ & $19.748\pm0.047$\\
\bottomrule
\end{tabular}
\end{table}
\FloatBarrier

\bibliographystyle{abbrvnat}
\bibliography{references}

@inproceedings{transformer,
 author = {Vaswani, Ashish and Shazeer, Noam and Parmar, Niki and Uszkoreit, Jakob and Jones, Llion and Gomez, Aidan N and Kaiser, {\L}ukasz and Polosukhin, Illia},
 booktitle = {Advances in Neural Information Processing Systems},
 editor = {I. Guyon and U. Von Luxburg and S. Bengio and H. Wallach and R. Fergus and S. Vishwanathan and R. Garnett},
 pages = {},
 publisher = {Curran Associates, Inc.},
 title = {Attention is All you Need},
 url = {https://proceedings.neurips.cc/paper_files/paper/2017/file/3f5ee243547dee91fbd053c1c4a845aa-Paper.pdf},
 volume = {30},
 year = {2017}
}

@misc{silu,
      title={Sigmoid-Weighted Linear Units for Neural Network Function Approximation in Reinforcement Learning}, 
      author={Stefan Elfwing and Eiji Uchibe and Kenji Doya},
      year={2017},
      eprint={1702.03118},
      archivePrefix={arXiv},
      primaryClass={cs.LG},
      url={https://arxiv.org/abs/1702.03118}, 
}

@inproceedings{rmsnorm,
 author = {Zhang, Biao and Sennrich, Rico},
 booktitle = {Advances in Neural Information Processing Systems},
 editor = {H. Wallach and H. Larochelle and A. Beygelzimer and F. d\textquotesingle Alch\'{e}-Buc and E. Fox and R. Garnett},
 pages = {},
 publisher = {Curran Associates, Inc.},
 title = {Root Mean Square Layer Normalization},
 url = {https://proceedings.neurips.cc/paper_files/paper/2019/file/1e8a19426224ca89e83cef47f1e7f53b-Paper.pdf},
 volume = {32},
 year = {2019}
}

@misc{rope,
      title={{RoFormer}: Enhanced Transformer with Rotary Position Embedding}, 
      author={Jianlin Su and Yu Lu and Shengfeng Pan and Ahmed Murtadha and Bo Wen and Yunfeng Liu},
      year={2023},
      eprint={2104.09864},
      archivePrefix={arXiv},
      primaryClass={cs.CL},
      url={https://arxiv.org/abs/2104.09864}, 
}

@misc{swiglu,
      title={{GLU} Variants Improve Transformer}, 
      author={Noam Shazeer},
      year={2020},
      eprint={2002.05202},
      archivePrefix={arXiv},
      primaryClass={cs.LG},
      url={https://arxiv.org/abs/2002.05202}, 
}

@inproceedings{adamw,
      title={Decoupled Weight Decay Regularization},
      author={Ilya Loshchilov and Frank Hutter},
      booktitle={International Conference on Learning Representations},
      year={2019},
      url={https://openreview.net/forum?id=Bkg6RiCqY7},
}

@inproceedings{fineweb,
      author = {Penedo, Guilherme and Kydl\'{\i}\v{c}ek, Hynek and Ben Allal, Loubna and Lozhkov, Anton and Mitchell, Margaret and Raffel, Colin and Von Werra, Leandro and Wolf, Thomas},
      booktitle = {Advances in Neural Information Processing Systems},
      doi = {10.52202/079017-0970},
      editor = {A. Globerson and L. Mackey and D. Belgrave and A. Fan and U. Paquet and J. Tomczak and C. Zhang},
      pages = {30811--30849},
      publisher = {Curran Associates, Inc.},
      title = {The {FineWeb} Datasets: Decanting the Web for the Finest Text Data at Scale},
      url = {https://proceedings.neurips.cc/paper_files/paper/2024/file/370df50ccfdf8bde18f8f9c2d9151bda-Paper-Datasets_and_Benchmarks_Track.pdf},
      volume = {37},
      year = {2024}
}

@inproceedings{gau,
      title = 	 {Transformer Quality in Linear Time},
      author =       {Hua, Weizhe and Dai, Zihang and Liu, Hanxiao and Le, Quoc},
      booktitle = 	 {Proceedings of the 39th International Conference on Machine Learning},
      pages = 	 {9099--9117},
      year = 	 {2022},
      editor = 	 {Chaudhuri, Kamalika and Jegelka, Stefanie and Song, Le and Szepesvari, Csaba and Niu, Gang and Sabato, Sivan},
      volume = 	 {162},
      series = 	 {Proceedings of Machine Learning Research},
      month = 	 {17--23 Jul},
      publisher =    {PMLR},
      url = 	 {https://proceedings.mlr.press/v162/hua22a.html}
    }

@inproceedings{mega,
      title={Mega: Moving Average Equipped Gated Attention},
      author={Xuezhe Ma and Chunting Zhou and Xiang Kong and Junxian He and Liangke Gui and Graham Neubig and Jonathan May and Luke Zettlemoyer},
      booktitle={The Eleventh International Conference on Learning Representations },
      year={2023},
      url={https://openreview.net/forum?id=qNLe3iq2El}
      }

@inproceedings{gatedattention,
      author = {Qiu, Zihan and Wang, Zekun and Zheng, Bo and Huang, Zeyu and Wen, Kaiyue and Yang, Songlin and Men, Rui and Yu, Le and Huang, Fei and Huang, Suozhi and Liu, Dayiheng and Zhou, Jingren and Lin, Junyang},
      booktitle = {Advances in Neural Information Processing Systems},
      doi = {10.52202/085713-3345},
      editor = {D. Belgrave and C. Zhang and H. Lin and R. Pascanu and P. Koniusz and M. Ghassemi and N. Chen},
      pages = {100092--100118},
      publisher = {Curran Associates, Inc.},
      title = {Gated Attention for Large Language Models: Non-linearity, Sparsity, and Attention-Sink-Free},
      url = {https://proceedings.neurips.cc/paper_files/paper/2025/file/904e89bb4e632e75fb47f093b620b257-Paper-Conference.pdf},
      volume = {38, Main Conference},
      year = {2025}
      }

@inproceedings{laser,
  title = 	 {{LASER}: Attention with Exponential Transformation},
  author =       {Duvvuri, Sai Surya and Dhillon, Inderjit S},
  booktitle = 	 {Proceedings of the 42nd International Conference on Machine Learning},
  pages = 	 {14902--14917},
  year = 	 {2025},
  editor = 	 {Singh, Aarti and Fazel, Maryam and Hsu, Daniel and Lacoste-Julien, Simon and Berkenkamp, Felix and Maharaj, Tegan and Wagstaff, Kiri and Zhu, Jerry},
  volume = 	 {267},
  series = 	 {Proceedings of Machine Learning Research},
  month = 	 {13--19 Jul},
  publisher =    {PMLR},
  url = 	 {https://proceedings.mlr.press/v267/duvvuri25a.html}
}

@misc{neuralattention,
      title={Neural Attention: Enhancing {QKV} Calculation in Self-Attention Mechanism with Neural Networks}, 
      author={Muhan Zhang},
      year={2023},
      eprint={2310.11398},
      archivePrefix={arXiv},
      primaryClass={cs.CL},
      url={https://arxiv.org/abs/2310.11398}, 
}

@inproceedings{hypernetwork,
      title={Attention as a Hypernetwork},
      author={Simon Schug and Seijin Kobayashi and Yassir Akram and Joao Sacramento and Razvan Pascanu},
      booktitle={The Thirteenth International Conference on Learning Representations},
      year={2025},
      url={https://openreview.net/forum?id=V4K9h1qNxE}
      }

@misc{gpt,
  author = {Radford, Alec and Narasimhan, Karthik and Salimans, Tim and Sutskever, Ilya},
  title = {Improving language understanding by generative pre-training},
  url = {https://cdn.openai.com/research-covers/language-unsupervised/language_understanding_paper.pdf},
  urldate = {2023-01-06},
  year = 2018
}

@inproceedings{benchvariance,
 author = {Bouthillier, Xavier and Delaunay, Pierre and Bronzi, Mirko and Trofimov, Assya and Nichyporuk, Brennan and Szeto, Justin and Mohammadi Sepahvand, Nazanin and Raff, Edward and Madan, Kanika and Voleti, Vikram and Ebrahimi Kahou, Samira and Michalski, Vincent and Arbel, Tal and Pal, Chris and Varoquaux, Gael and Vincent, Pascal},
 booktitle = {Proceedings of Machine Learning and Systems},
 editor = {A. Smola and A. Dimakis and I. Stoica},
 pages = {747--769},
 title = {Accounting for Variance in Machine Learning Benchmarks},
 url = {https://proceedings.mlsys.org/paper_files/paper/2021/file/0184b0cd3cfb185989f858a1d9f5c1eb-Paper.pdf},
 volume = {3},
 year = {2021}
}

@inproceedings{glu,
  title = 	 {Language Modeling with Gated Convolutional Networks},
  author =       {Yann N. Dauphin and Angela Fan and Michael Auli and David Grangier},
  booktitle = 	 {Proceedings of the 34th International Conference on Machine Learning},
  pages = 	 {933--941},
  year = 	 {2017},
  editor = 	 {Precup, Doina and Teh, Yee Whye},
  volume = 	 {70},
  series = 	 {Proceedings of Machine Learning Research},
  month = 	 {06--11 Aug},
  publisher =    {PMLR},
  url = 	 {https://proceedings.mlr.press/v70/dauphin17a.html}
}

@techreport{gpt2,
  author = {Radford, Alec and Wu, Jeffrey and Child, Rewon and Luan, David and Amodei, Dario and Sutskever, Ilya},
  title = {Language Models Are Unsupervised Multitask Learners},
  institution = {OpenAI},
  year = {2019},
  url = {https://cdn.openai.com/better-language-models/language_models_are_unsupervised_multitask_learners.pdf}
}

@inproceedings{statsguide,
    title = "The Hitchhiker{'}s Guide to Testing Statistical Significance in Natural Language Processing",
    author = "Dror, Rotem  and
      Baumer, Gili  and
      Shlomov, Segev  and
      Reichart, Roi",
    editor = "Gurevych, Iryna  and
      Miyao, Yusuke",
    booktitle = "Proceedings of the 56th Annual Meeting of the {A}ssociation for {C}omputational {L}inguistics (Volume 1: Long Papers)",
    month = jul,
    year = "2018",
    address = "Melbourne, Australia",
    publisher = "Association for Computational Linguistics",
    url = "https://aclanthology.org/P18-1128/",
    doi = "10.18653/v1/P18-1128",
    pages = "1383--1392"
}

@misc{swish,
      title={Searching for Activation Functions}, 
      author={Prajit Ramachandran and Barret Zoph and Quoc V. Le},
      year={2017},
      eprint={1710.05941},
      archivePrefix={arXiv},
      primaryClass={cs.NE},
      url={https://arxiv.org/abs/1710.05941}, 
}

@misc{wikitext,
      title={Pointer Sentinel Mixture Models}, 
      author={Stephen Merity and Caiming Xiong and James Bradbury and Richard Socher},
      year={2016},
      eprint={1609.07843},
      archivePrefix={arXiv},
      primaryClass={cs.CL},
      url={https://arxiv.org/abs/1609.07843}, 
}

@misc{gluattention,
      title={{GLU} Attention Improve Transformer}, 
      author={Zehao Wang},
      year={2025},
      eprint={2507.00022},
      archivePrefix={arXiv},
      primaryClass={cs.LG},
      url={https://arxiv.org/abs/2507.00022}, 
}
\end{document}